Autonomous Navigation Series

# Object Detection and Ranging for Autonomous Navigation of Mobile Robots

*Md Ziaul Haque Zim[a,b]\* and Nimai Chandra Das[b]*

[a]Department of Computer Science and Engineering, Daffodil International University, Dhaka, Bangladesh
[b]Department of Automatic Control Systems, Saint Petersburg Electrotechnical University "LETI", Saint Petersburg, Russia

ABSTRACT

In the recent decade, electronic technology gets advanced day by day the methodologies too should update. For the purpose of ranging various methods such Radio Detection and Ranging (RADAR), Light Detection and Ranging (LIDAR) and Sonic Navigation and Ranging (SONAR) etc. are used. Later, by adapting the earlier technologies and further modifying the purposes of detection and ranging in navigation, the technology of Sonic Detection and Ranging (SODAR) is used in modern robotics. The SODAR can be defined as a child of SONAR and also a twin of Echo sounder. The echo-sounder is used only for ranging. But the SODAR use the low-frequency wave of 33 kHz to measure the underwater depth and also to detect the objects below the water medium. So, this work comprises the designing of a system to evaluate the Object Detection and Ranging for Autonomous Navigation of Mobile Robots.

## Motivation

With the advent of Industry 4.0, factory floors will be more connected, with decentralized communications and decision-making being performed on the factory floor. Autonomous robots and smart assembly lines shall become an integral part of the smart factory. Autonomous robots can deliver parts from the storage area and the workstation and load or unload goods from vehicles. In this research, we are attempting to teach autonomous machines about environment by adding a sensor adaptation layer to the general architecture of an autonomous robotic platform.

Many of the techniques which are used for robot perception and continuously adapting these technologies to the environment floor would help us produce better quality products at an economical price. Adapting this continuously could become tedious and costly, so we propose a new module in the general autonomous vehicle architecture with which we can augment data at the sensor output to ensure the mobile robot follows its environment rules. While several autonomous vehicles and robots rely on complex sensors, the following research questions arise:

a. Is it possible to build a low-cost AUGV [Autonomous Unmanned Ground Vehicle]?
b. Can the AUGV exhibit reliable autonomous decision making to move from one place to another by understanding its environment?
c. Can the AUGV detect obstacles and process this information to calculate the best path to reach its destination?
d. Can the system easily adapt to new environments/conditions?

## 1. Introduction

Robotics has helped humans greatly in automating many activities in the factory. Robots have revolutionized multiple industrial processes to mass produce products several decades ago, but these robots are hardly more than versatile machines running a complex, but fixed program. In general, manufacturing robots do not exhibit autonomous intelligence. Except for basic control flow, they are mostly unaware of their environment and limited sensory input is used during operation. Therefore, great care is taken so that the environment of the robot is as predictable as possible, e.g., by using fences to lock uncertainties out of the robot's workspace. This lack of autonomy is one of the major obstacles which must be overcome to allow robots to become mobile in a prior known environment. In many cases, robots are controlled manually to move from source to destination. However, several studies have been carried out on autonomous robots leading to a whole panoply of potential applications. Autonomous navigation of robots on the factory floor will bootstrap its capabilities, navigation is a complex task that relies on developing an internal representation of space, grounded by recognizable landmarks and robust visual processing, that can simultaneously support continuous self-localization and a representation of the destination. When a robot is deployed in an environment such as a factory, it is usually not feasible to equip the robot with an accurate model of that environment in prior. Therefore, the robot will first need to create a model of its world. For a mobile robot, this model generally needs to comprises a map that allows it to localize itself and plan a collision-free path according to its assignment.

\* *Corresponding author.*
E-mail address: ziaul15-1133@diu.edu.bd

## 2. Strategy for Robot Navigation

Figure 1 depicts the block diagram of the global strategy for constructing a local occupancy map of indoor environments. As the operating range of the ultrasonic sensor is known, this sets the size of the area covered by the sensor, which will be directly updated in the local map at each instant time t. This area establishes the size of an initial local map previously provided to the robot. On the other hand, before robot starts to move, the ultrasonic sensor is used for detecting objects in front of him. To do this, the ultrasonic sensor carries out a "sweep" in the range −90° to 90° with respect to X axis of the robot (see Figure 2a). The local map also uses the initial robot location, due to location of the objects are given with respect to the ultrasonic sensor. The path-planning module computes an initial path for robot navigation. The potential field as technique is used to plan the best trajectory for the robot. In real time, robot odometrical location is obtained from the encoders for updating the map and continuously constructing the global map of the robot scene. Both global and local maps store the probability of occupancy around the robot during navigation.

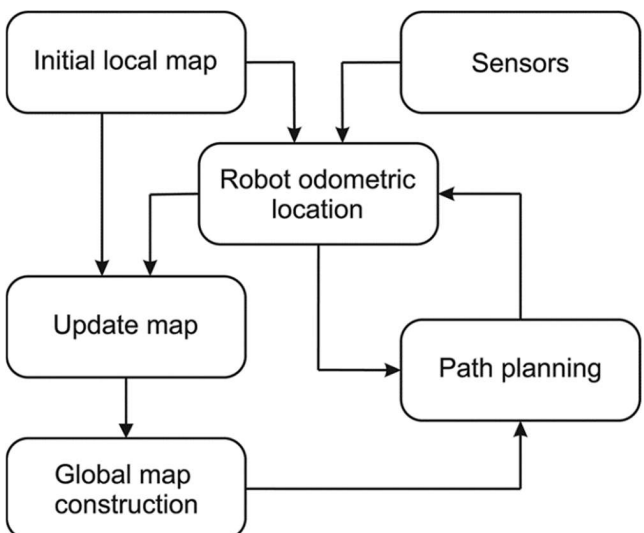

**Fig. 1 - Global strategy for constructing an occupancy map during indoor navigation.**

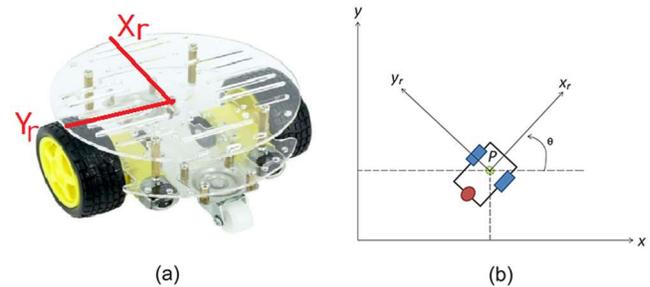

**Fig. 2 - (a) Reference axis of the mobile robot (b) Graphical representation of the robot in the local and global reference axis, P represents the reference point of the position.**

Every update of the global map is stored in a file internally on the robot. The map construction consists in dividing the environment in small uniform cells, which will be labelled as occupied or free in accordance with the ultrasonic measures. An intensive calibration strategy is required with the aim of obtaining an accurate digital representation of the real scene. The robot navigates from a predefined initial position to a goal position; then, the algorithm ends when the robot reaches such predefined goal position.

## 3. The Sensor

An ultrasonic sensor is an instrument that measures the distance to an object using ultrasonic sound waves. It uses a transducer to send and receive ultrasonic pulses that relay back information about an object's proximity. High-frequency sound waves reflect from boundaries to produce distinct echo patterns. Its work by sending out a sound wave at a frequency above the range of human hearing. The transducer of the sensor acts as a microphone to receive and send the ultrasonic sound.

The working principle of this module is simple. It sends an ultrasonic pulse out at 40kHz which travels through the air and if there is an obstacle or object, it will bounce back to the sensor. By calculating the travel time and the speed of sound, the distance can be calculated. Ultrasonic sensors are a great solution for the detection of clear objects. For liquid level measurement, applications that use infrared sensors, for instance, struggle with this particular use case because of target translucence. For presence detection, ultrasonic sensors detect objects regardless of the color, surface, or material (unless the material is very soft like wool, as it would absorb sound.)

33To detect transparent and other items where optical technologies may fail, ultrasonic sensors are a reliable choice.

## 4. Example Code

If you want to detect movement you want to detect changes in the echo delay. So, what you do is setup your timer to consistently ping at a fixed rate then keep a low pass filtered running average of the echo delay and look at the differential of this (the rate of change).

```
float g = 0.9f
float avg_time = max_time
float avg_dt = 0.0f
float dt_hysteresis = 0.01f
for each 50ms
  if no echo for 10 iterations
    avg_time = max_time
    continue

  float echo_time = sensor.read()

  if avg_time == max_time
    avg_time = echo_time
    avg_dt = 0.0f
  else
    float prev_avg_time = avg_time;
    avg_time = avg_time * g + (1.0f - g) * echo_time
    avg_dt = avg_dt * g + (1.0f - g) * (avg_echo_time - prev_avg_time);

    if avg_dt < -dt_hysteresis
      // Some one approaching
    else if avg_dt > +dt_hysteresis
      // Some one leaving
  end for
```

**Explanation** g is a filter coefficient between 0 and 1.0, the higher it is the more "inert" your filter will be. It will be slower to react but be highly resistant to noise. If g is low the filter will react quickly but also more susceptible to noise, to put it simply. You can think of it as a low-pass filter where g implicitly controls the cut-off frequency. Normal values are in the range 0.7 to 0.95.

Every 50 ms you send out a ping and look for the echo. If you don't get an echo for 10 iterations or so, simply reset the avg_time variable so that you can start fresh the next time you get an echo and not be affected by what happened to be stored there since the last echo.

If you get an echo you low pass filter it with avg_time = avg_time * g + (1.0f - g) * echo_time and then look at the lowpass filtered change in the lowpass filtered time value avg_dt = avg_dt * g + (1.0 - g) * (avg_echo_time - prev_avg_time). If this change is larger than +- the hysteresis threshold dt_hysteresis you then deduce that someone is approaching or moving away from the sensor. You can also consider looking directly at avg_echo_time - prev_avg_time the extra lowpass filtering is simply to get rid of noise which is always present when we differentiate a real signal.

## 5. Calculations

Time of Flights (ToF) is the measurement of the time taken by an object, particle or wave to travel a distance through a medium. This information can then be used to establish a time standard, as a way to measure velocity or path length, or as a way to learn about the particle or medium's properties.

$$T = (2 \cdot u_y)/g$$
$$T = (2 U \sin\theta)/g$$

The time of flight of a projectile motion is the time from when the object is projected to the time it reaches the surface. As we discussed previously, T depends on the initial velocity magnitude and the angle of the projectile.

## 6. Conclusion

In this paper, we presented a novel approach of Object Detection and Ranging for Autonomous Navigation of Mobile Robots. We contribute a novel model for autonomous navigation, obstacle avoidance and object detection.

REFERENCES

ROS.org | Powering the world's robots,
  https://www.ros.org/ (accessed 21 May 2021).
OpenCV: Epipolar Geometry,
  https://docs.opencv.org/3.4/da/de9/tutorial_py_epipolar_geometry.html (accessed 20 June 2021)
Arduino: Ultrasonic Sensor
  Ultrasonic Sensor HC-SR04 with Arduino Tutorial - Arduino Project Hub (accessed 20 June 2021).
usb_cam - ROS Wiki,
  http://wiki.ros.org/usb_cam (accessed 21 June 2021).
stereo_image_proc - ROS Wiki,
  http://wiki.ros.org/stereo_image_proc (accessed 20 June 2021).